\documentclass[lettersize,journal,twoside]{IEEEtran}
\usepackage{amsmath,amsfonts}
\usepackage[vlined,ruled,linesnumbered]{algorithm2e}
\usepackage{array}
\usepackage[caption=false,font=normalsize,labelfont=sf,textfont=sf]{subfig}
\usepackage{textcomp}
\usepackage{stfloats}
\usepackage{url}
\usepackage{verbatim}
\usepackage{graphicx}
\usepackage{cite}

\usepackage{comment}
\usepackage{siunitx}
\usepackage{relsize}
\usepackage{ifthen}
\usepackage[colorinlistoftodos]{todonotes}
\usepackage[vlined,ruled,linesnumbered]{algorithm2e}
\usepackage{graphics} 
\usepackage{rotating}
\usepackage{color}
\usepackage{enumerate}
\usepackage[T1]{fontenc}
\usepackage{psfrag}
\usepackage{epsfig} 
\usepackage{booktabs}
\usepackage{graphicx,url}
\usepackage{multirow}
\usepackage{array}
\usepackage{latexsym}
\usepackage{amsfonts}
\usepackage{amsmath}
\usepackage{amssymb}
\usepackage{mathtools}
\usepackage{xstring}
\usepackage[noend]{algorithmic}
\usepackage{multirow}
\usepackage{xcolor}
\usepackage{prettyref}
\usepackage{flexisym}
\usepackage{bigdelim}
\usepackage{breqn} 
\usepackage{listings}

\usepackage{enumitem}
\usepackage{xspace}
\usepackage{bm}
\graphicspath{{./figures/}}
\usepackage{tikz}
\usetikzlibrary{matrix,calc}
\usepackage{tabularx}

\usepackage{mdwlist}

\let\itemize\undefined
\makecompactlist{itemize}{stditemize}

\usepackage{algorithm2e}
\usepackage{float}
\usepackage{booktabs}
\usepackage{adjustbox}

\usepackage{caption} 

\usepackage{placeins}
 
\usepackage{fancyhdr}
\usepackage{etoolbox}

\newrefformat{prob}{Problem\,\ref{#1}}
\newrefformat{def}{Definition\,\ref{#1}}
\newrefformat{sec}{Section\,\ref{#1}}
\newrefformat{sub}{Section\,\ref{#1}}
\newrefformat{prop}{Proposition\,\ref{#1}}
\newrefformat{app}{Appendix\,\ref{#1}}
\newrefformat{alg}{Algorithm\,\ref{#1}}
\newrefformat{cor}{Corollary\,\ref{#1}}
\newrefformat{thm}{Theorem\,\ref{#1}}
\newrefformat{lem}{Lemma\,\ref{#1}}
\newrefformat{fig}{Fig.\,\ref{#1}}
\newrefformat{tab}{Table\,\ref{#1}}

\newtheorem{theorem}{Theorem}

\newtheorem{remark}[theorem]{Remark}

\newcommand{\bdmath}{\begin{dmath}}
\newcommand{\edmath}{\end{dmath}}
\newcommand{\beq}{\begin{equation}}
\newcommand{\eeq}{\end{equation}}
\newcommand{\bdm}{\begin{displaymath}}
\newcommand{\edm}{\end{displaymath}}
\newcommand{\bea}{\begin{eqnarray}}
\newcommand{\eea}{\end{eqnarray}}
\newcommand{\beal}{\beq \begin{array}{ll}}
\newcommand{\eeal}{\end{array} \eeq}
\newcommand{\beas}{\begin{eqnarray*}}
\newcommand{\eeas}{\end{eqnarray*}}
\newcommand{\ba}{\begin{array}}
\newcommand{\ea}{\end{array}}
\newcommand{\bit}{\begin{itemize}}
\newcommand{\eit}{\end{itemize}}
\newcommand{\ben}{\begin{enumerate}}
\newcommand{\een}{\end{enumerate}}

\newcommand{\calD}{{\cal D}}

\newcommand{\calP}{{\cal P}}

\newcommand{\calR}{{\cal R}}

\newcommand{\calT}{{\rv{N_d}}}

\newcommand{\calW}{{\cal W}}

\newcommand{\hide}[1]{}

\newcommand{\hiddenText}{{\color{gray} hidden text.}}
\newcommand{\hideWithText}[1]{\hiddenText}

\DeclareMathOperator*{\argmax}{arg\,max}

\newcommand{\scenario}[1]{{\smaller \sf#1}\xspace}

\newcommand{\blue}[1]{{\color{blue}#1}}

\newcommand{\linkToPdf}[1]{\href{#1}{\blue{(pdf)}}}
\newcommand{\linkToPpt}[1]{\href{#1}{\blue{(ppt)}}}
\newcommand{\linkToCode}[1]{\href{#1}{\blue{(code)}}}
\newcommand{\linkToWeb}[1]{\href{#1}{\blue{(web)}}}
\newcommand{\linkToVideo}[1]{\href{#1}{\blue{(video)}}}
\newcommand{\linkToMedia}[1]{\href{#1}{\blue{(media)}}}
\newcommand{\award}[1]{\xspace} 

\newcommand{\bmat}{\left[ \begin{array}}
\newcommand{\emat}{\end{array}\right]}

\newcommand{\nameshort}{\scenario{GPC}}
\newcommand{\gpcrank}{\scenario{GPC-RANK}}
\newcommand{\gpcrankopt}{\scenario{GPC-RANK+OPT}}
\newcommand{\gpcopt}{\scenario{GPC-OPT}}
\newcommand{\expert}{\text{expert}}
\newcommand{\explore}{\text{explore}}

\newcommand{\vision}{\text{vision}}

\usepackage{xcolor}

\newcommand{\rv}[1]{#1}

\begin{document}

\title{Inference-Time Enhancement of Generative Robot Policies via Predictive World Modeling }


\author{
Han Qi, 
Haocheng Yin, 
Aris Zhu,
Yilun Du,
and Heng Yang
\thanks{
Manuscript received: October 19, 2025; Revised: January 31, 2026; Accepted: February 28, 2026.
}
\thanks{This paper was recommended for publication by Editor Cosimo Della Santina upon evaluation of the Associate Editor and Reviewers comments.}

\thanks{Han Qi, Aris Zhu, Yilun Du, and Heng Yang are with Harvard University, Cambridge, MA, USA. Haocheng Yin is with Georgia Institute of Technology, Atlanta, GA, USA. Han Qi and Heng Yang acknowledge funding from Office of Naval Research grant N00014-25-1-2322.}

\thanks{Corresponding author: Han Qi (\texttt{hqi@g.harvard.edu}).}

\thanks{Digital Object Identifier (DOI): see top of this page.}

}

\markboth{IEEE ROBOTICS AND AUTOMATION LETTERS. PREPRINT VERSION. ACCEPTED FEBRUARY, 2026}{Qi \MakeLowercase{\textit{et al.}}: Inference-Time Enhancement of Generative Robot Policies via Predictive World Modeling}

\maketitle





\begin{figure*}[t]
  \centering
  \vspace{2mm}
  \begin{minipage}{\textwidth}
    \centering
        \includegraphics[width=0.9\textwidth]{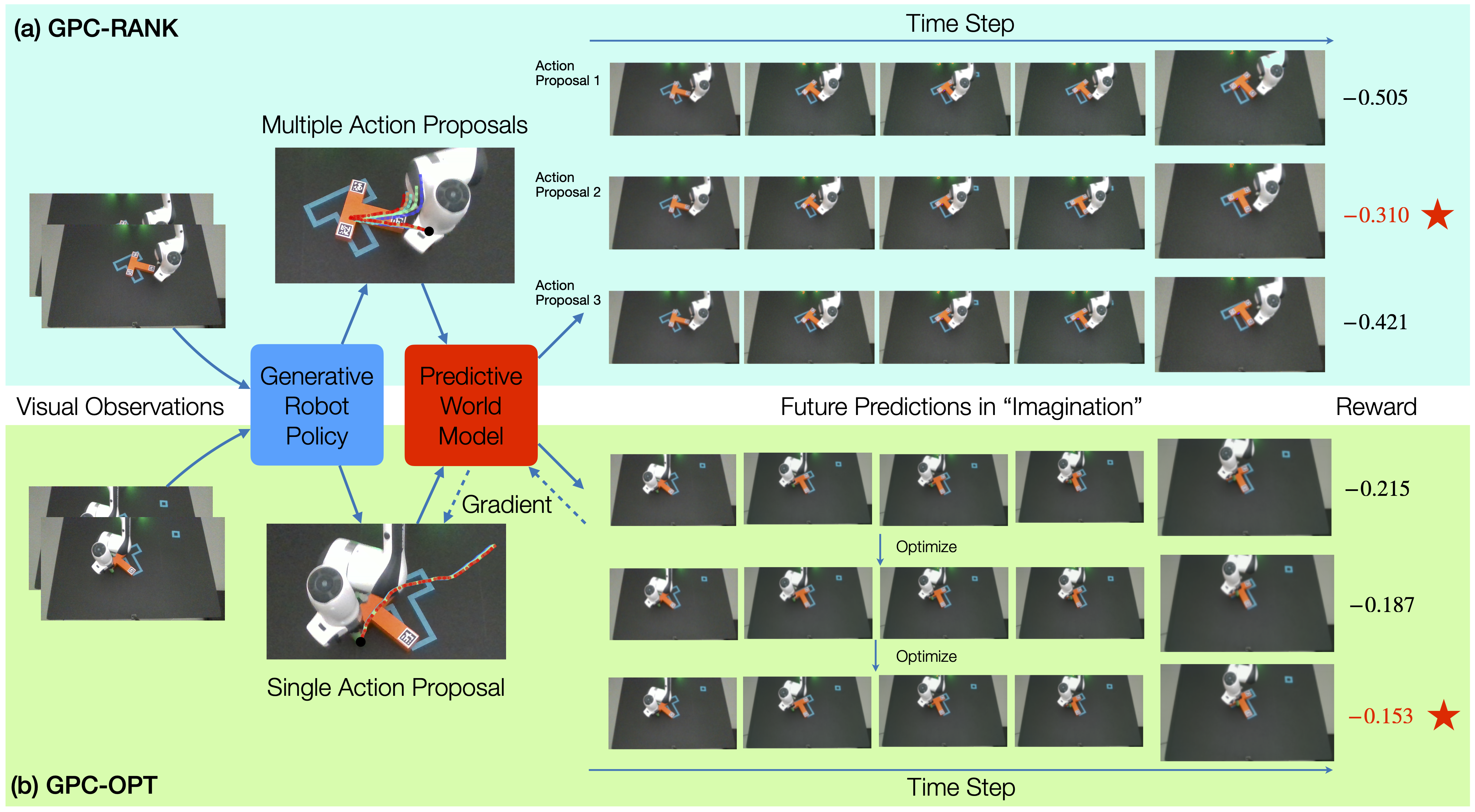}
    \caption{
    \textsc{\rv{Generative Predictive Control (GPC).}} 
    (a) \textbf{GPC-RANK:} The generative policy proposes multiple action sequences that are evaluated in imagination using the predictive world model; the action with the highest predicted reward is selected. 
    (b) \textbf{GPC-OPT:} A single action proposal is refined via gradient-based optimization through the world model to maximize the predicted reward. 
    Together, these strategies enable inference-time enhancement of pretrained behavior cloning policies by combining generative sampling with predictive foresight.
    }
    \label{fig:overview}
  \end{minipage}
  \vspace{-5mm}
\end{figure*}


\begin{abstract}

We present \emph{generative predictive control} (\nameshort), a framework for \emph{inference-time} enhancement of pretrained behavior-cloning policies. Rather than retraining or fine-tuning, \nameshort augments a frozen diffusion policy at deployment by coupling it with a \emph{predictive world model}. Concretely, we train an action-conditioned world model on expert demonstrations and random exploration rollouts to forecast the consequences of action proposals produced by the diffusion policy, then perform lightweight online planning that ranks and refines these proposals via model-based look-ahead. This combination of a generative prior with predictive foresight enables test-time adaptation. Across diverse robotic manipulation tasks---state- and vision-based, in simulation and on real hardware---\nameshort consistently outperforms standard behavior cloning and compares favorably to other inference-time adaptation baselines.

\end{abstract}


\section{Introduction}
\label{sec:introduction}

\IEEEPARstart{B}{ehavior} cloning (BC) with generative models has become a central paradigm for robot policy learning, enabling robots to imitate expert demonstrations and generalize across diverse manipulation tasks~\cite{chi2023diffusion,urain2024deep,firoozi2023foundation,kim2024openvla}. At its core, generative control \emph{looks back}, grounding decisions in previously observed expert behavior.

Despite their success, BC policies are often brittle at deployment. Lacking explicit mechanisms for test-time correction or recovery, small deviations from the training distribution can compound over time and degrade performance~\cite{li2024evaluating}. By contrast, model predictive control (MPC) \emph{looks ahead}: it evaluates candidate actions by simulating their future consequences under a predictive dynamics model, enabling online adaptation. While MPC-style planning has demonstrated robustness across robotics and control, it typically relies on carefully engineered models and objectives, making direct integration with modern generative policies challenging.

This paper is motivated by the question: \emph{Can we endow pretrained, frozen BC policies with test-time adaptability by incorporating MPC-style foresight through learned world models—without retraining or fine-tuning the policy itself?} \rv{Inspired by how humans combine retrospective experience with prospective mental simulation, we seek a lightweight inference-time framework that unifies these two modes of reasoning, combining BC’s generative flexibility with predictive foresight in a form that is adaptive and interpretable.}

\textbf{Contribution.} We propose \emph{generative predictive control} (\nameshort), a framework that strengthens pretrained diffusion-based BC policies at inference time by coupling them with an action-conditioned predictive world model for online planning (Fig.~\ref{fig:overview}). \nameshort consists of three components:
\begin{itemize}
    \item \textbf{Generative policy training.} From expert demonstrations, we train a diffusion-based policy that generates short-horizon \emph{action chunks} conditioned on past observations, providing a generative prior over plausible behaviors.
    
    \item \textbf{Predictive world modeling.} We learn an action-conditioned world model that forecasts future observations given candidate action chunks. Training solely on demonstrations yields a narrow model that captures only expert behavior; we therefore augment training with simple random exploration data to enrich the learned dynamics and enable corrective predictions. For state-based tasks, we use MLPs; for vision-based tasks, we employ conditional video diffusion models~\cite{ho2022video,alonso2024diffusion}.
    
    \item \textbf{Online planning.} At inference time, \nameshort enhances the frozen policy using lightweight planning strategies. \gpcrank samples multiple action proposals, unrolls them through the world model, and selects the one with the highest predicted reward. \gpcopt treats a policy sample as a warm start and refines it via gradient-based optimization through the world model. These strategies can be combined, and rewards can be either learned from demonstrations or provided zero-shot by a vision--language model (VLM).
\end{itemize}

Across simulated and real-world manipulation tasks, \nameshort consistently outperforms pure behavior cloning and compares favorably to other inference-time adaptation methods, demonstrating that predictive world modeling and lightweight planning provide an effective recipe for enhancing generative robot policies at deployment.

\rv{
\textbf{Novelty.} While GPC is related to inference-time planning methods that enhance frozen policies via imagined rollouts in learned world models~\cite{huang2025ladi,hafner2023mastering}, it is distinguished by combining a diffusion policy with an explicit, image-space diffusion world model, introducing a \emph{frozen-noise inference mechanism} for stable gradient-based optimization, unifying proposal ranking and refinement, and enabling vision--language models to act as direct reward surrogates.
}


\section{Related Work}
\label{sec:related-work}

\textbf{Generative modeling for robotics}. 
A large body of recent work has explored the integration of generative models into robotics~\cite{urain2024deep}. 
Generative models have been widely applied to represent policies~\cite{chi2023diffusion, qi2024control, ankile2024imitation, lee2024behavior, qi2025compose}, often trained on large collections of task demonstrations. 
They have also been leveraged to generate additional data~\cite{wang2023robogen, wang2023gensim, chen2023genaug} and to model world dynamics~\cite{liu2022structdiffusion, byravan2017se3, yang2023learning}. 
More recently, several works have used generative models directly for planning~\cite{yang2024diffusion, carvalho2023motion, luo2024potential}, training them over trajectories of states and actions and exploiting the generation process for action optimization. 
In contrast, our \emph{generative predictive control} (\nameshort) framework targets \emph{inference-time planning}: rather than retraining or modifying the generative policy, we leave it fixed and enhance its execution through a predictive world model. 
This modular design decouples policy learning from world model learning, allowing them to be trained independently and even from different datasets.

\textbf{Visual world modeling for predictive policy learning}.
Learning predictive visual models of the world has a long history~\cite{oh2015action, ha2018world, oprea2020review}.  Recently, video generative models have emerged as a powerful tool for modeling the physical world~\cite{du2024learning, yang2023learning, liang2024dreamitate, ko2023learning}. Such models have been used to initialize policies~\cite{du2024learning, ko2023learning, bharadhwaj2024gen2act}, as interactive simulators~\cite{yang2023learning, liang2024dreamitate, bar2024navigation}, and integrated with downstream planning~\cite{du2023video,wang2024language}. Some recent work (e.g., IRIS \cite{micheli2022transformers}, Dreamer-v3 \cite{hafner2023mastering}, TDMPC-2 \cite{hansen2023td}) proposes to train learning-based policies in the imagined world model to solve simple game tasks or simulated control tasks. Besides, several other work \cite{zhoudino, huang2025ladi} focus on MPC-style optimization with predictive world modeling in \emph{latent space} during inference time. 

Different from these works, our approach employs a video-based world model as an \emph{explicit action-conditioned dynamics model} that directly predicts future observations. \rv{This design enables inference-time planning that strengthens pretrained policies by allowing direct, \emph{interpretable} evaluation of predicted outcomes, facilitating robust manipulation in both simulation and the real world as presented in our experiments.}

\vspace{-1mm}


\setlength{\intextsep}{0pt}

\vspace{2mm}
\begin{algorithm}
\caption{Generative Predictive Control (\nameshort)}
\label{alg:gpc}

\KwIn{Expert action demonstrations $\calD^P_\expert = \{I_t^i \leftrightarrow a^i_{t:t+T}\}_{i=1}^{N_\expert}$; 
Expert trajectories and random exploration trajectories $\calD^W_\expert \cup \calD^W_\explore = \{ (I_t^i,a^i_{t:t+T}) \leftrightarrow I^i_{t+1:t+T+1} \}_{i=1}^{N_\expert + N_\explore}$; 
Reward model $\calR$; Positive integers $K$, $M$}

\rv{
\tcp{Instantiation: $M=0$ yields ranking-only GPC-RANK; 
$K=1$ and $M>0$ yields optimization-only GPC-OPT.}}

$\calP(\cdot) \leftarrow \textsc{BehaviorCloning}(\calD^P_{N_\expert})$\;

$\calW(\cdot) \leftarrow \textsc{DynamicsLearning}(\calD^W_\expert \cup \calD^W_\explore)$\;

\For{$k=1,\dots,K$}{
    $a^{(k)}_{t:t+T} \sim \calP(I_t)$\;
    Set $\hat{a}^{(0)}_{t:t+T} = a^{(k)}_{t:t+T}$\;
    \For{$\ell=1,\dots,M$}{
        $\hat{a}^{(\ell)}_{t:t+T} = \hat{a}^{(\ell-1)}_{t:t+T} + \eta^{(\ell)} 
        \nabla_{a_{t:t+T}} \calR(\calW(I_t, \hat{a}_{t:t+T}^{(\ell-1)}))$\;
    }
    Set $a^{(k)}_{t:t+T} = \hat{a}^{(M)}_{t:t+T}$\;
}
Find $k_\star \in \argmax_{k=1,\dots,K} \ \calR(\calW(I_t, a_{t:t+T}^{(k)}))$\;
\Return $a_{t:t+T}^{(k_\star)}$\;
\end{algorithm}
\vspace{-2mm}
\vspace{-1mm}
\section{Overview of Generative Predictive Control}
\label{sec:gpc}

Let $I_t$ be the information vector summarizing the state of the robot and the environment up to time step $t$. For state-based robot control, $I_t := x_{t-H:t}$ is the history of (low-dimensional) state of the robot and its environment (e.g., poses); for vision-based robot control, $I_t := o_{t-H:t}$ is the history of (high-dimensional) visual observations where each $o_t$ is a single (or multi-view) image(s). Let $a_t$ be the robot action at time $t$ and $a_{t:t+T}$ be an \emph{action chunk} of length $T+1$. Our goal is to design a policy that decides $a_{t:t+T}$ based on $I_t$ to solve certain tasks (e.g., described by language instructions). 
We formalize the three modules of \nameshort.

\textbf{Generative policy training}. 
We ask humans to teleoperate robots to solve tasks, generating expert demonstrations that are segmented into clips of state-action pairs \rv{with sliding windows}, forming a dataset \rv{$\calD_{\expert}^P = \{ I_t^{i}, a_{t:t+T}^{i} \}_{i=1}^{N_\expert}$}. Policy learning then reduces to supervised learning with input $I_t$ and output $a_{t:t+T}$.
In \nameshort, we adopt the diffusion policy learning framework \cite{chi2023diffusion}, where 
$I_t$ is fed into a network that parametrizes the score function of $p(a_{t:t+T} |I_t)$. Gaussian noise vectors are gradually denoised into expert action chunks following \cite{ho2020denoising}. This yields a \emph{stochastic} policy network 
$\calP(\cdot)$ that samples action chunks $a_{t:t+T}$ given $I_t$. We refer to each sampled action chunk as an \emph{action proposal}.

\rv{
In implementation, we follow the standard Diffusion Policy temporal abstraction, using an observation horizon $H=2$, a prediction horizon $T=16$, and an action horizon of 9, with control executed in a receding-horizon manner. We resize the images to height of 96. The generative policy $\calP(\cdot)$ is trained using a DDPM formulation~\cite{chi2023diffusion}, employing a ResNet18 visual encoder for vision-based policy and a UNet diffusion backbone to model the action distribution. Training is performed for $300$ epochs using AdamW with learning rate $10^{-4}$ and weight decay $10^{-6}$, and the policy uses $100$ diffusion denoising steps during inference.

}

\textbf{Predictive world modeling}. 
The stochastic nature of $\calP(\cdot)$ raises a key question: which of the many action proposals should be selected? The world model addresses this by predicting the future outcomes of each proposal. To train it, we use pairs of $(I_t, a_{t:t+T})$ and $I_{t+1:t+T+1}$ from the expert dataset \rv{$\calD_\expert^W = \{(I^i_t, a^i_{t:t+T}), I_{t+1:t+T+1}^i \}_{i=1}^{N_\expert}$}, already used in policy training. However, as shown in \S\ref{sec:experiments}, training solely on $\calD_\expert^W$ leads to limited predictive ability. While more expert data could help, it can be costly. Instead, we collect an additional exploration dataset \rv{$\calD^W_\explore = \{(I^i_t, a^i_{t:t+T}), I_{t+1:t+T+1}^i \}_{i=1}^{N_\explore}$}, where humans (or other controllers) randomly perturb the system without solving tasks. This approach, inspired from ``system identification'' with ``sufficient excitation'' in control theory \cite{lennart1999system}, enriches the dynamics. We combine both datasets as $\calD^W := \calD^W_\expert \cup \calD^W_\explore$ to train the world model. Architectural details are in \S\ref{sec:world-model}; for now, we assume access to a model $\calW(\cdot)$ that predicts $I_{t+1:t+T+1}$ from $(I_t, a_{t:t+T})$.

\textbf{Online planning}. We now formalize the two online planning algorithms, namely \gpcrank and \gpcopt, that combine the policy $\calP(\cdot)$ and the world model $\calW(\cdot)$. We assume a reward model $\calR(\cdot): I_{t+1:t+T+1} \rightarrow r_{t}$ is available that predicts the reward given the future state information, e.g., a small neural network that is separately trained or VLMs capable of zero-shot selection of the most promising future state. A trained reward predictor is appropriate in scenarios where the reward can be explicitly defined—particularly in numerical terms—and it is differentiable, enabling gradient-based action optimization. However, some tasks involve rewards that are difficult or even infeasible to specify. In such cases, VLMs can serve as surrogate reward predictors by directly selecting the most suitable action proposal, given the predicted outcomes and task description as input. This approach introduces greater flexibility to our \nameshort method, extending its applicability to a broader range of tasks.

\begin{enumerate}[label=(\roman*)]
    \item \gpcrank is based on the intuition to pick the action proposal with highest reward. Formally, we sample $K$ action proposals from the policy
    \bea \label{eq:sample}
    (a_{t:t+T}^{(1)},\dots, a_{t:t+T}^{(k)}, \dots, a_{t:t+T}^{(K)}) \sim \calP(I_t),
    \eea
    pass them through the world model $\calW(\cdot)$, and select the one with highest reward:
    \bea \label{eq:rank}
    \pi(I_t) = a_{t:t+T}^{(k_\star)}, \quad k_\star \in \argmax_{k=1,\dots,K} \ \calR(\calW(I_t, a_{t:t+T}^{(k)})),
    \eea
    where we used the notation $\pi(\cdot)$ to denote the online policy in contrast to the offline policy $\calP(\cdot)$. \rv{A key advantage of \gpcrank is its simplicity: it is easily parallelizable, requires no hyperparameter tuning, and applies broadly across tasks and reward types, including non-differentiable and VLM-based rewards.}

    \item \gpcopt directly solves the reward maximization problem given the world model:
    \bea \label{eq:max-reward}
    \max_{a_{t:t+T}} \calR(\calW(I_t, a_{t:t+T})),
    \eea
    treating the action chunk as decision variables. However, \eqref{eq:max-reward} is often highly nonconvex due to the complexity of $\calW(\cdot)$ and $\calR(\cdot)$. To address this, we leverage the pretrained $\calP(\cdot)$ to \emph{warm-start} optimization. We sample $\hat{a}_{t:t+T}$ from $\calP(I_t)$ and perform:
    
    \begin{align} \label{eq:ascent}
    \hat{a}_{t:t+T}^{(0)} 
        &:= \hat{a}_{t:t+T} \sim \mathcal{P}(I_t), \nonumber \\
    \hat{a}_{t:t+T}^{(\ell)} 
        &= \hat{a}_{t:t+T}^{(\ell-1)} 
           + \eta^{(\ell)} \nabla_{a_{t:t+T}} 
             \mathcal{R}\!\big(\mathcal{W}(I_t, \hat{a}_{t:t+T}^{(\ell-1)})\big),
    \end{align}
    for $\ell = 1, \dots, M$, with step sizes $\eta^{(\ell)}>0$.  
    The final optimized action chunk is:
    \begin{align} \label{eq:gpc-opt}
        \pi(I_t) = \hat{a}_{t:t+T}^{(M)}.
    \end{align}

   Gradients are obtained via automatic differentiation, and updates can be performed with optimizers like ADAM. \gpcopt resembles the classical single shooting method~\cite{diehl2011numerical}. Unlike \gpcrank, which evaluates the world model $K$ times in parallel, \gpcopt evaluates it $M$ times sequentially. \rv{In contrast, \gpcopt enables continuous action refinement by performing gradient-based optimization from diffusion-policy warm starts, allowing it to improve beyond sampled proposals. Under appropriate optimization step sizes and iteration counts, this refinement is particularly effective for tasks with reliable numerical rewards (see \S\ref{sec:experiments}).
}
\end{enumerate}

\gpcrank and \gpcopt can be combined by sampling $K$ action proposals from $\calP(\cdot)$ and applying reward maximization~\eqref{eq:ascent} to each, yielding $K$ optimized action chunks. We select the one with the highest reward. \gpcrankopt effectively solves~\eqref{eq:max-reward} from multiple initializations~\cite{sharony2024learning}, requiring $K \times M$ world model evaluations.
A summary of \nameshort is provided in Alg.~\ref{alg:gpc}. The world model is crucial for ranking and optimizing action proposals. While early works such as deep visual foresight~\cite{finn2017deep} explored learned world models, we next show that modern diffusion models now enable (near) physics-accurate visual predictions previously unattainable.



\section{World Model Learning}
\label{sec:world-model}
\begin{figure}[t]
\vspace{3mm}
\begin{minipage}{0.45\textwidth}
    \centering
    \includegraphics[width=\linewidth]{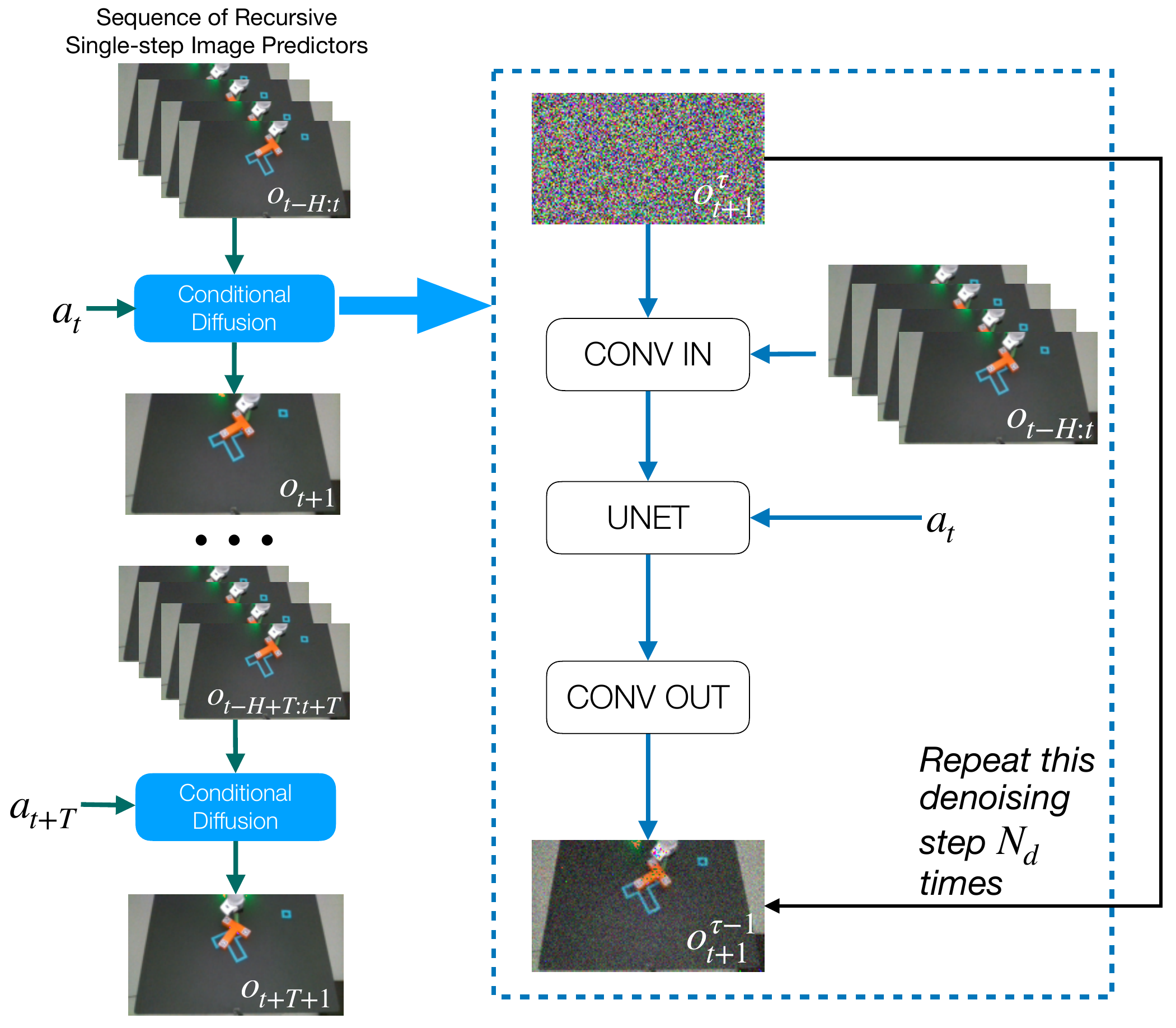}
    \caption{\textsc{Diffusion-based visual world modeling.} \rv{\textbf{[Left]} Recursive single-step prediction produces multi-step futures.
    \textbf{[Right]} Each single-step predictor is a conditioned diffusion model, where a UNet iteratively denoises a noisy image sample conditioned on the observation history and action for $N_d$ steps.}}

    \label{fig:visual-world-model}
    \vspace{-4mm}
\end{minipage}
\end{figure}

The world model takes as input $(I_t, a_{t:t+T})$ and predicts $I_{t+1:t+T+1}$. When $I_t$ represents a low-dimensional state, an MLP can be used to learn the dynamics. However, state-based world modeling relies on accurate state estimation, which is feasible in controlled lab environments with infrastructure such as AprilTags~\cite{olson2011apriltag} or motion capture systems, but can be challenging in open or unstructured environments. Therefore, it is desirable to learn \emph{visual dynamics} directly, i.e., to build a visual world model.

\textbf{Diffusion-based visual world modeling}. Motivated by the success of using diffusion models for image generation, we design a visual world model based on conditional diffusion. Recall that the input to the world model is $I_t = o_{t-H:t}$ (the sequence of past images) and $a_{t:t+T}$, and the output is $I_{t+1:t+T+1}:= o_{t+1:t+T+1}$ (a sequence of future images). We design the visual world model as a recursive application of single-step image predictors, which reads
\vspace{-1mm}
\bea \label{eq:single-step-predictor}
o_{t+1} = f_\vision(o_{t-H:t}, a_t),
\eea 
\vspace{-1mm}
where $f_\vision$ is a conditional diffusion model. The unique property of diffusion is that $o_{t+1}$ is not generated in a single feedforward step, but rather through a sequence of denoising steps. In particular, let $o_{t+1}^{\calT}$ be drawn from a white Gaussian noise distribution, then $f_\vision$ proceeds as:
\bea \label{eq:image-diffusion}
o_{t+1}^{\tau-1} = D_{\phi}(o_{t+1}^\tau, \tau, o_{t-H:t},a_t), \tau = \calT, \dots, 1,
\eea
where $\calT$ is the total number of denoising steps and the output of $f_\vision$ is $o_{t+1} = o_{t+1}^0$. In~\eqref{eq:image-diffusion}, $\tau$ is the denoising step index and $D_{\phi}$---the ``denoiser''---is a neural network with weights $\phi$. We use the same architecture for $D_\phi$ as~\cite{alonso2024diffusionworldmodelingvisual}, containing convolutions, action embedding, and a U-Net (Fig.~\ref{fig:visual-world-model}). $D_\phi$ is trained by adding random noises to the clean images and then predicting the noise.

\textbf{Two-phase training}.
To improve the accuracy and consistency of visual world modeling, we train it in two phases. In phase one, we train only the single-step image predictor~\eqref{eq:single-step-predictor}, i.e., with supervision only from a single image $o_{t+1}$. In phase two, we recursively apply the single-step predictor~\eqref{eq:single-step-predictor} for $T$ times to obtain a sequence of future images $o_{t+1:t+T+1}$, and jointly supervise them with groundtruth images. 
We use observation horizon $H=4$ in the visual world modeling, and $\calT=3$ diffusion steps. \rv{In practice, we choose the EDM-based diffusion model for high quality output with less denoising steps \cite{alonso2024diffusion}.}

\begin{remark}[Freeze the Noise]
    The single-step image predictor $f_{\vision}$ in~\eqref{eq:single-step-predictor} is inherently stochastic due to the random initialization of the noise $o_{t+1}^{\calT}$. While such stochasticity is beneficial for generative diversity, our objective is control, where we seek to isolate the effect of actions on future outcomes rather than noise. We therefore fix $o_{t+1}^{\calT}=0$ at inference time, making the world model deterministic and producing the most likely future prediction. Without freezing the noise, \gpcopt\ fails, as stochastic gradients destabilize the reward optimization in~\eqref{eq:ascent}.
\end{remark}


\section{Experiments}
\label{sec:experiments}


\renewcommand{\arraystretch}{1.5} 

\begin{table*}[h]
\vspace{2mm}
\centering
\begin{adjustbox}{width=\textwidth}
\begin{tabular}{@{} l c ccc c ccc  c @{}}
\toprule
 & Behavior Cloning 
 & \multicolumn{3}{c}{GPC-RANK} 
 & GPC-OPT 
 & \multicolumn{3}{c}{GPC-RANK+OPT} 
 & With GT Simulator \\
\cmidrule(lr){2-2} \cmidrule(lr){3-5} \cmidrule(lr){6-6} \cmidrule(lr){7-9} \cmidrule(lr){10-10}
 & $(K{=}1, M{=}0)$ 
 & $(K{=}100, M{=}0)$ & $(K{=}1000, M{=}0)$ & $(K{=}5000, M{=}0)$ 
 & $(K{=}1, M{=}30)$ 
 & $(K{=}10, M{=}30)$ & $(K{=}20, M{=}30)$ & $(K{=}30, M{=}20)$ 
 &  \\ 
\midrule
Score 
 & 0.812 
 & 0.898 & 0.932 & 0.934
 & 0.886 
 & 0.912 & 0.914 & 0.891
 & 0.952 \\ 
\bottomrule
\end{tabular}
\end{adjustbox}
\vspace{-1mm}
\caption{\textsc{State-based Planar Pushing (Push-T) in Simulation}. Scores report IoU averaged over 100 evaluation seeds. 
\rv{This table presents an ablation over sampling (i.e., number of action proposals $K$ from $\mathcal{P}(\cdot)$) and optimization (i.e., number of gradient steps $M$), illustrating the trade-offs and showing all \nameshort variants outperform pure behavior cloning.}}

\label{table:state-base-gpc}
\vspace{2mm}
\end{table*}

\begin{figure*}[t]
    \centering
    \vspace{-3mm}
    \begin{minipage}[t][\height][t]{0.49\textwidth}
        \centering
        \includegraphics[width=\linewidth]{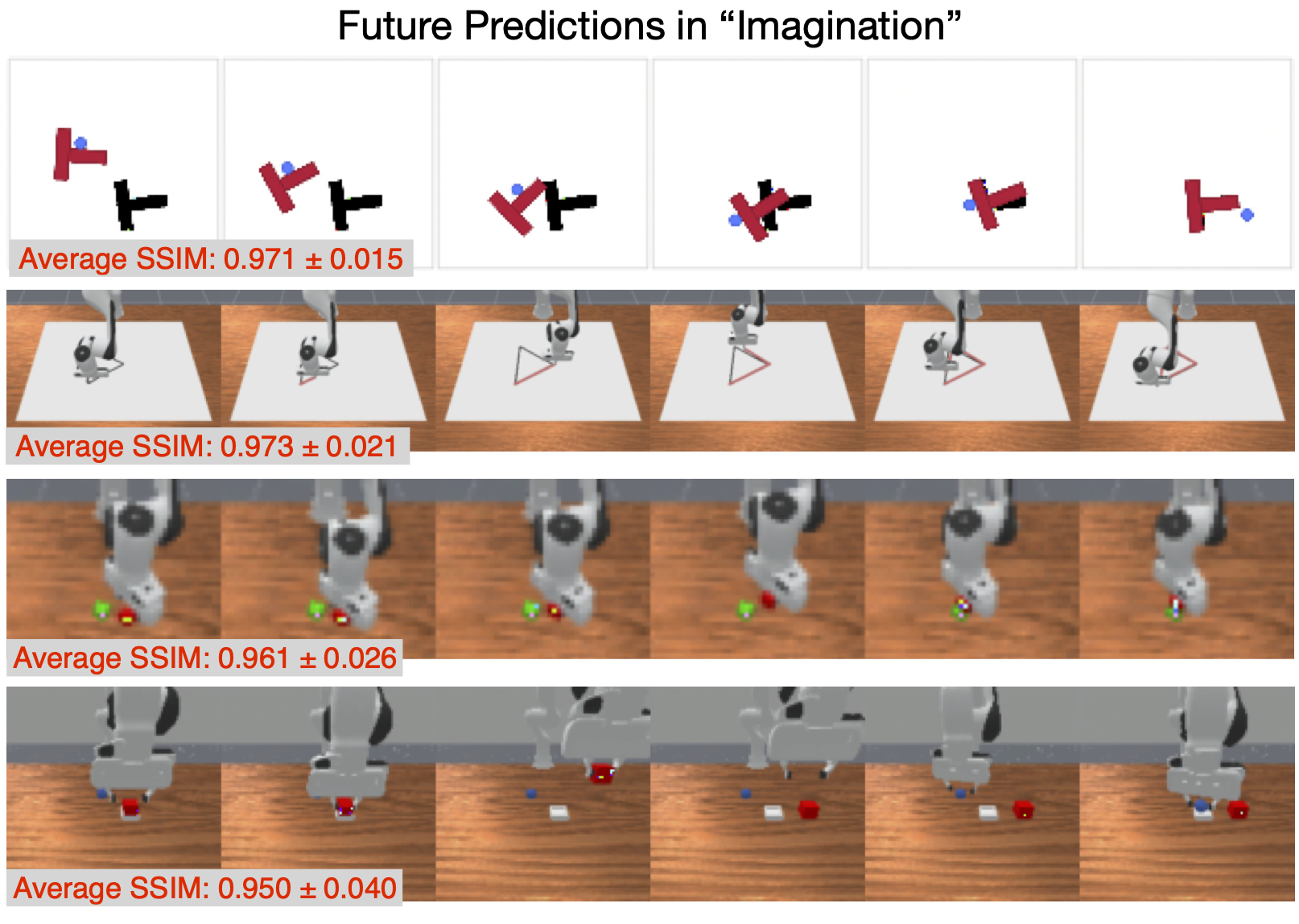}
        \vspace{-5mm}
        \caption{\textsc{World model predictions in GPC for vision-based simulation tasks.} 
        All images shown are model-predicted future observations, sampled from intermediate steps along the rollout horizon. 
        \rv{We report the average structural similarity index (SSIM) between predicted and ground-truth frames over the full evaluation horizon ($\approx$ $250$ frames), averaged across $5$ evaluation seeds.}
        }
        \vspace{7mm}
        \label{fig:simulation-imagination}
    \end{minipage}
    \hfill
    \begin{minipage}[t][\height][t]{0.49\textwidth}
        \centering
        \vspace{-14\baselineskip}

        \renewcommand{\arraystretch}{4.2} 

        \begin{adjustbox}{width=\linewidth}
        \begin{tabular}{@{} l c c c c c @{}} 
        \toprule
         & Diff. Policy & V-GPS & LaDi-WM & DreamerV3 & \textbf{GPC-RANK} \\
        \cmidrule(lr){2-2} \cmidrule(lr){3-3} \cmidrule(lr){4-4} \cmidrule(lr){5-5} \cmidrule(lr){6-6}
        Push-T & 0.642 & 0.620 & 0.683 & 0.649 & \textbf{0.739} \\[-1mm]
        Triangle Drawing & 0.724 & 0.593 & 0.680 & 0.732 & \textbf{0.767} (0.761) \\[-1mm]
        Block Stacking & 0.912 & 0.972 & 0.933 & 0.983 & \textbf{0.989} (0.973) \\[-1mm]
        Cube \& Sphere Swapping & 0.680 & 0.650 & 0.410 & 0.700 & \textbf{0.730} \\
        \bottomrule
        \end{tabular}
        \end{adjustbox}

        \vspace{0mm}
        \captionof{table}{\textsc{Scores for four vision-based tasks in simulation.} 
        For \gpcrank, rewards are obtained from a learned predictor (\rv{$K=100$ for learned reward predictor}) by default, 
        with parentheses showing results when using a VLM (\rv{$K=10$ for VLM reward predictor}).}
        \label{table:simulation_score}
        \vspace{-20mm}
    \end{minipage}
     \vspace{-12mm}
\end{figure*}

We evaluate \nameshort on (1) a state-based planar pushing task, (2) four vision-based simulation tasks, and (3) two real-world manipulation tasks. In all cases, \nameshort consistently outperforms the behavior cloning baseline, highlighting its effectiveness as an \emph{inference-time enhancement}. We further provide ablations and comparisons to illustrate: (i) the influence of $K$ and $M$ on performance, and (ii) how \nameshort compares with other baselines designed to strengthen behavior cloning at inference time.

\subsection{State-based Planar Pushing in Simulation}
\label{sec:push-T-sim-state}

\rv{In this section, $I_t$ is the history of (low-dimensional) state.} We study the planar pushing task with the goal of pushing an object from an initial pose to a specified target pose, where the groundtruth pose of the object is available through a simulator. We firstly train a state-based diffusion policy $\calP(\cdot)$ \cite{chi2023diffusion} from which we can sample multiple \emph{action proposals}, as explained in \S\ref{sec:gpc}. Then with both expert demonstrations and random exploration data, we utilize MLP networks to construct the state-based world model predicting the outcome states and define the reward with a \emph{registration loss} between the target and predicted object states. \rv{The registration loss is defined as the $\ell_2$ distance between transformed object vertices under predicted and target poses..}

In Table~\ref{table:state-base-gpc}, we present results for \nameshort and the pure behavior cloning baseline ($K=1$ and $M=0$). Clearly, all \gpcrank, \gpcopt, and \gpcrankopt variants outperform pure behavior cloning. \rv{Notably, the best-performing GPC variant in Table~\ref{table:state-base-gpc} approaches the performance of planning based on a pretrained behavior cloning policy with a \emph{groundtruth simulator} (i.e., a groundtruth world model)}

\begin{figure*}
    \centering
    \vspace{2mm}
    \includegraphics[width=\linewidth]{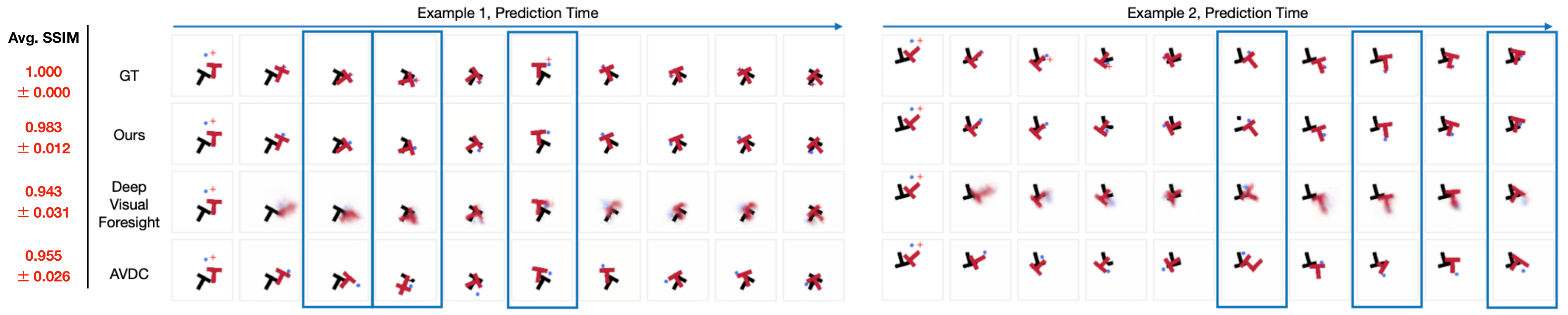}
    \caption{\textsc{Comparison of different visual world models.} \rv{We report the average SSIM between each method's prediction and ground-truth frames over $10$ uniformly sampled prediction from the full evaluation, averaged across $100$ samples for each method. Two representative sets of predicted frames are shown, with regions of interest highlighted in blue for comparison.} \label{fig:compare-world-model}}
    \vspace{-2mm}
\end{figure*}

\subsection{Vision-based Tasks in Simulation}
\label{sec:push-T-sim-vision}


\begin{table*}[h]
    \setlength{\heavyrulewidth}{0.08em}
    \setlength{\lightrulewidth}{0.05em}
    \centering
    \renewcommand{\arraystretch}{1.5}

    \begin{adjustbox}{width=0.98\textwidth}
    \begin{tabular}{@{} l c c c c c @{}}  
        \toprule
        \multicolumn{6}{c}{Impact of $K$ and $M$ on \nameshort} \\
        \midrule
        Method & Behavior Cloning\ ($K{=}1, M{=}1$)  & \gpcrank\ ($K{=}50, M{=}0$) & \gpcrank\ ($K{=}100, M{=}0$) & \gpcopt\ ($K{=}1, M{=}25$) & \gpcrankopt\ ($K{=}10, M{=}25$) \\
        \cmidrule(lr){2-2}\cmidrule(lr){3-3}\cmidrule(lr){4-4}\cmidrule(lr){5-5}\cmidrule(lr){6-6}
        Score (IoU) & 0.642 & 0.698 & 0.739 & 0.791 & 0.882 \\

        Timing (s) & 0.457 & 5.835 & 11.745 & 39.061 & 374.102 \\
        \bottomrule
    \end{tabular}
    \end{adjustbox}


    \caption{\textsc{Ablation Study for Vision-based Planar Pushing (Push-T) in Simulation.}
    Score is measured by the IoU metric averaged over 100 evaluation seeds. \rv{We also provide wall‑clock timing of one decision cycle for pure behavior cloning and \nameshort with different scale of $K$ and $M$.}
    $K$ is the number of action proposals; $M$ is the number of gradient steps.}
    \label{table:vision-base-gpc}
    \vspace{-4mm}
\end{table*}

\rv{In this section, $I_t$ is a sequence of (high-dimensional) visual observations.} We test on four vision-based simulation tasks named as vision-based Push-T, Triangle Drawing, Block Stacking, and Cube \& Sphere Swapping. We train a vision-based diffusion policy $\calP(\cdot)$~\cite{chi2023diffusion} using ResNet18 and UNet. For the world model, as in \S\ref{sec:world-model}, we use a diffusion model to build a single-step image predictor and recursively apply it $T$ times to form the world model $\calW(\cdot)$. 
The training data for the world model includes random explorations.

\textbf{Quality of visual world modeling.} 
\rv{Fig.~\ref{fig:simulation-imagination} shows that our world model generates visually realistic future predictions with accurate object interactions. 
Prediction quality is quantified using the structural similarity index (SSIM) between predicted and ground-truth frames, where SSIM is bounded above by $1.0$ and higher values indicate closer visual correspondence.} Besides, we compare the world model against two baselines: deep visual foresight~\cite{finn2017deep}, which uses CNNs and LSTMs for prediction,\footnote{\url{https://github.com/Xiaohui9607/physical_interaction_video_prediction_pytorch}} and AVDC~\cite{du2024learning}, a video diffusion model originally conditioned on language that we adapt to robot actions.\footnote{\url{https://github.com/flow-diffusion/AVDC}} 
Unlike our recursive design, AVDC predicts multiple future steps jointly. 
As shown in Fig.~\ref{fig:compare-world-model}, diffusion-based approaches (ours and AVDC) outperform deep visual foresight, with our model producing predictions more closely aligned with groundtruth.

\textbf{Reward.} Unlike state-based planar pushing, where future states are structured, image-based predictions make reward computation difficult. We adopt two strategies: (1) when a numerical reward can be defined (e.g., registration loss in Push-T or cube distance in Block Stacking), we train a ResNet18+MLP reward predictor $\mathcal{R}(\cdot)$ that is differentiable and suitable for online planning; (2) optionally, we can leverage a VLM (e.g., ChatGPT-4o \cite{openai2024chatgpt4o}) to select among action proposals in a zero-shot manner, prompting it with predicted images and the task description to identify the most promising proposal. \rv{For each proposal, we extract the final predicted frames of prediction (resolution 96) and provide these images to the VLM along with a fixed, task-specific prompt (e.g., ``selecting the image that best satisfies the task objective''). The VLM is queried once per decision step to select the most promising candidate. We set VLM temperature to $0.2$.}

\textbf{Planning performance.} Table~\ref{table:simulation_score} reports the results of \gpcrank alongside the pure behavior cloning baseline (Diffusion Policy~\cite{chi2023diffusion}), with scores averaged over $50$ evaluation seeds. We also compare against three inference-time enhancement methods: (1) LaDi-WM~\cite{huang2025ladi}, which uses a latent diffusion-based world model built on pretrained visual features for optimization in imagined \emph{latent states}; (2) V-GPS~\cite{nakamoto2024steering}, which improves generalist policies by re-ranking action proposals with a value function learned via offline reinforcement learning; and (3) DreamerV3~\cite{hafner2023mastering}, a latent-space world model used for action selection. \rv{For fairness, all baselines are trained on the same data and share the same pretrained diffusion policy. During evaluation, we use $100$ action candidates for all baselines.}
\rv{Using either a learned reward predictor $\mathcal{R}(\cdot)$ with 100 candidates or a VLM with 10 candidates, \gpcrank achieves strong performance across all four vision-based tasks, with the best-performing \gpcrank variant attaining the highest overall results.}


\begin{figure}[t]
    \centering
    \includegraphics[width=0.4\textwidth]{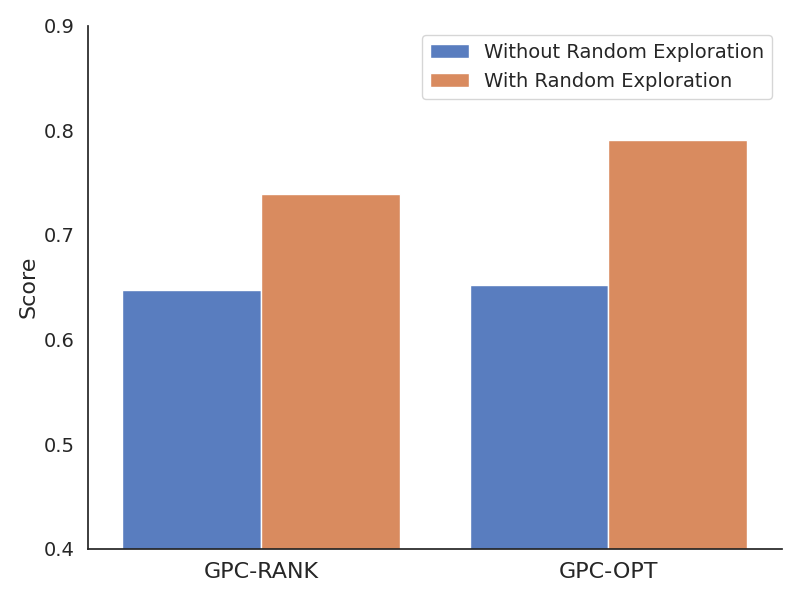}
    \vspace{-2mm}
    \caption{\textsc{Importance of random exploration in world model learning (vision-based push-T).}}
    \label{fig:vision-base-compare-only-expert}
    \vspace{-3mm}
\end{figure}

\textbf{Ablations in planar pushing.} Using the Push-T task, we analyze the impact of $K$ and $M$, compare against additional MPC-style baselines without diffusion-policy warm-start, and study the role of random exploration. 

\begin{itemize}
    \item \emph{Impact of $K$ and $M$.} Table~\ref{table:vision-base-gpc} (Top) reports how varying $K$ and $M$ affects performance. 
Because image-based experiments are more computationally demanding than state-based ones, we restrict $K$ to the scale of hundreds. 
The results show that (a) \gpcrank improves performance by $\sim\!10\%$ over the behavior cloning baseline; (b) \gpcopt yields a $\sim\!15\%$ gain; and (c) \gpcrankopt achieves up to $\sim\!25\%$.  

    \item \emph{Importance of combining the generative prior with predictive foresight.} Planning-only methods without a generative policy prior, including model predictive path integral (MPPI), cross-entropy method (CEM), and pure gradient ascent~\cite{zhoudino}, achieve substantially lower performance on vision-based Push-T, with success rates below $0.2$. In contrast, \nameshort, which combines diffusion-based behavior cloning with predictive world modeling, consistently achieves much higher performance, highlighting the importance of integrating generative priors with inference-time planning.

    \item \emph{Importance of random exploration.} Fig.~\ref{fig:vision-base-compare-only-expert} compares \gpcrank and \gpcopt using world models trained with and without random exploration. 
Introducing exploration improves performance by about $10\%$, underscoring its importance for accurate world modeling.
\end{itemize}

\rv{In summary, \gpcrank and \gpcopt offer complementary benefits. \gpcrank emphasizes simplicity, parallelism, and broad applicability to diverse tasks, while \gpcopt enables continuous refinement for tasks with reliable numerical rewards; their combination (\gpcrankopt) explores the full potential of \nameshort at increased inference-time cost.}


\begin{figure*}[h]
    \vspace{5mm}
    \begin{minipage}{\textwidth}
        \centering
        \begin{tabular}{c}
            \hspace{-4mm}
            \begin{minipage}{\textwidth}
                \centering
                \includegraphics[width=\columnwidth]{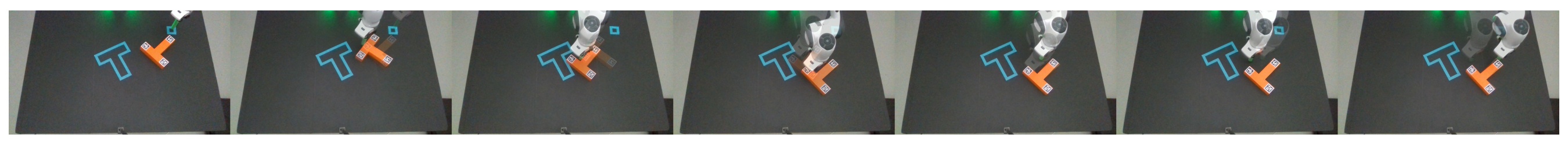}
            \end{minipage}
            \\
            \hspace{-4mm}
            \begin{minipage}{\textwidth}
                \centering
                \includegraphics[width=\columnwidth]{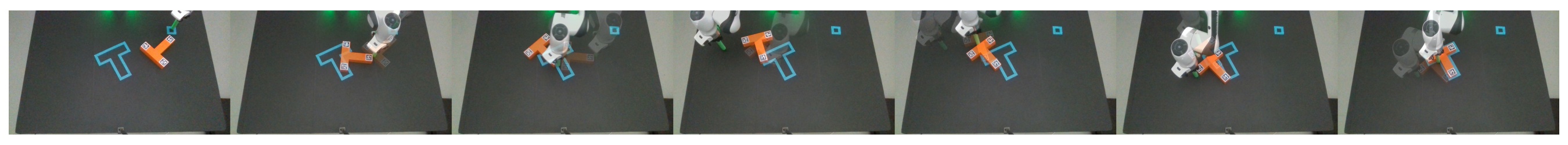}
            \end{minipage}
            \\
            \hspace{-4mm}
            \begin{minipage}{\textwidth}
                \centering
                \includegraphics[width=\columnwidth]{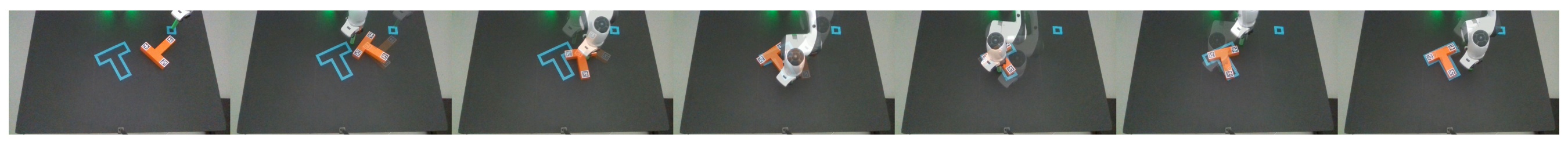}
            \end{minipage}

            \\ 
        \end{tabular}
    \end{minipage}
    \vspace{-2mm}
    \caption{\textsc{Real-world tests for push-T.} Top row shows trajectories of baseline model ($K=1, M=0$), middle row shows trajectories of \gpcrank ($K=10, M=0$), and last row shows trajectories of \gpcopt ($K=0, M=25$). 
    \label{fig:fig-real_world_main_paper_push_t}}
    \vspace{-5mm}
\end{figure*}


\vspace{-3mm}

\begin{figure*}[h]
    \begin{minipage}{\textwidth}
        \centering
        \begin{tabular}{c}
            \hspace{-4mm}
            \begin{minipage}{\textwidth}
                \centering
                \includegraphics[width=\columnwidth]{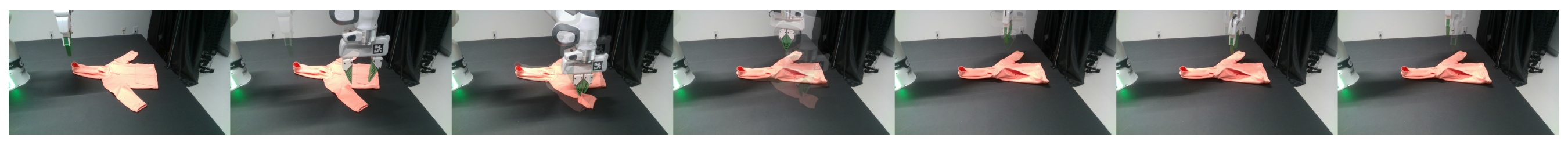}
            \end{minipage}
            \\
            \hspace{-4mm}
            \begin{minipage}{\textwidth}
                \centering
                \includegraphics[width=\columnwidth]{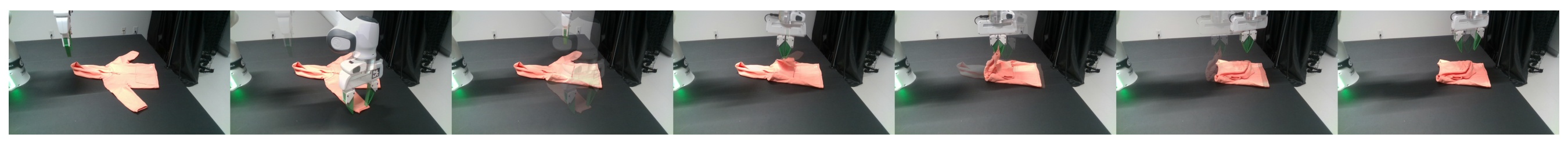}
            \end{minipage}
            \\

        \end{tabular}
    \end{minipage}
    \vspace{-2mm}
    \caption{\textsc{Real-world tests for clothes folding.} Top row shows trajectories of baseline model ($K=1, M=0$), second row shows trajectories of \gpcrank ($K=10, M=0$). 
    \label{fig:fig-real_world_main_paper_fold_clothes}}
    \vspace{-4mm}
\end{figure*}

\begin{figure}[t]
    \centering
    \includegraphics[width=0.4\textwidth]{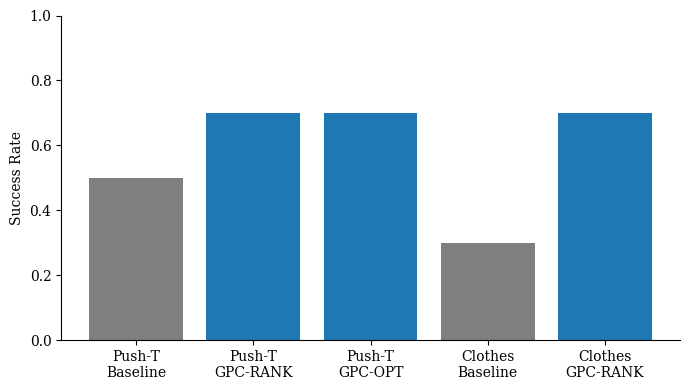}
    \vspace{-2mm}
    \caption{\textsc{\rv{Real-World Success Rates}}}
    \label{fig:fig-real_world_success_rates}
    \vspace{-5mm}
\end{figure}

\vspace{2mm}

\subsection{Real-world Vision-based Tasks}
\label{sec:real-world}

\rv{In this section, $I_t$ is sequences of (high-dimensional) visual observations.}  We test \nameshort on real-world tasks, involving Push-T and clothes folding. For the Push-T tasks, rewards are computed using a learned reward predictor composed of a ResNet and MLP, trained with registration losses derived from expert demonstrations' state information using AprilTags~\cite{olson2011apriltag}. However for cloth folding, where states are harder to define, we design a progressing reward: starting at $0$ for spread clothes and increasing with each successful grasp-and-move operation. Normalized progressing rewards for expert demonstrations are used to train the reward predictor for clothes folding. \rv{Fig.~\ref{fig:fig-real_world_success_rates} reports real-world success rates for two real-world tasks, plain Push-T and clothes folding (see trajectories in Fig.~\ref{fig:fig-real_world_main_paper_push_t} and Fig.~\ref{fig:fig-real_world_main_paper_fold_clothes}), over 10 trials.}

Despite the challenging nature of these real-world experiments—where dynamics are more complex due to collisions and tasks like cloth folding involve non-rigid objects and require multiple steps—\nameshort still operates effectively. \rv{Overall, while low-level state information is optionally used to train the reward predictor, both the policy and the world model are fully vision-based, and all components operate solely on visual observations during inference.
}


\section{Conclusion}

\rv{We presented \nameshort, a generative predictive control framework that enhances frozen behavior cloning policies at inference time by integrating predictive world modeling with lightweight online planning. By combining diffusion-based generative priors, action-conditioned visual world models, and flexible reward specification, \nameshort enables robust test-time adaptation without retraining, achieving strong performance across simulated and real-world manipulation tasks.}


\section{Limitation and Future Work}
\label{sec:limitation}

A limitation of \nameshort is its inference-time cost, since diffusion-based world-model rollouts dominate runtime ($\approx$90--95\%). While sufficient for our manipulation tasks ($\approx$3~s per decision cycle in real-world \gpcrank), improving efficiency remains an important direction, e.g., via diffusion distillation, faster solvers, or hardware acceleration.


\bibliographystyle{IEEEtran}
\bibliography{IEEEabrv,refs}
\appendix

\renewcommand{\theequation}{A\arabic{equation}} 
\setcounter{equation}{0} 
\renewcommand{\thefigure}{A\arabic{figure}} 
\setcounter{figure}{0} 
\renewcommand{\thetheorem}{A\arabic{theorem}} 
\setcounter{theorem}{0} 


\rv{
\section{Implementation Details and Reproducibility}
\label{app:app_detail}

\subsection{Tasks and Datasets}
For simulation, we collect expert demonstrations for policy learning and random perturbation trajectories for world-model training:
\begin{itemize}
    \item \textbf{Push-T:} 500 expert demos, $6{\times}500$ perturbations, 7 long-horizon demos.
    \item \textbf{Triangle Drawing:} 30 expert, 100 perturbations.
    \item \textbf{Block Stacking:} 50 expert, 100 perturbations.
    \item \textbf{Cube \& Sphere:} 100 expert, 100 perturbations.
\end{itemize}
For real-world tasks, we collect 100 expert demonstrations per task and 5 random-play videos (each several minutes).
}

\end{document}